\documentclass[runningheads]{llncs}
\usepackage[T1]{fontenc}
\usepackage{graphicx}
\usepackage{amsfonts}
\usepackage{amssymb}
\usepackage{multirow}
\usepackage{hyperref}
\hypersetup{colorlinks}
\usepackage[misc]{ifsym}
\begin{document}
%
\title{Contrastive Domain Disentanglement for Generalizable Medical Image Segmentation}
\titlerunning{Contrastive Domain Disentanglement}
%
\author{Ran Gu\inst{1, 2} \and
Jiangshan Lu\inst{1} \and
Jingyang Zhang\inst{3} \and 
Wenhui Lei\inst{2, 4} \and
Xiaofan Zhang\inst{2, 4} \and
Guotai Wang\inst{1, 2, }$^{\textrm{\Letter}}$ \and
Shaoting Zhang\inst{1, 2}
}
\authorrunning{R. Gu et al.}
%
\institute{School of Mechanical and Electrical Engineering, University of Electronic Science and Technology of China, Chengdu, China
\and
Shanghai AI Lab, Shanghai, China
\and
School of Biomedical Engineering, ShanghaiTech University, Shanghai, China.
\and
Qingyuan Institute, Shanghai Jiao Tong University, Shanghai, China \\
\email{guotai.wang@uestc.edu.cn}}
%
\maketitle              
\begin{abstract}
Efficiently utilizing discriminative features is crucial for convolutional neural networks to achieve remarkable performance in medical image segmentation and is also important for model generalization across multiple domains, where letting model recognize domain-specific and domain-invariant information among multi-site datasets is a reasonable strategy for domain generalization. Unfortunately, most of the recent disentangle networks are not directly adaptable to unseen-domain datasets because of the limitations of offered data distribution. To tackle this deficiency, we propose Contrastive Domain Disentangle (CDD) network for generalizable medical image segmentation. We first introduce a disentangle network to decompose medical images into an anatomical representation factor and a modality representation factor. Then, a style contrastive loss is proposed to encourage the modality representations from the same domain to distribute as close as possible while different domains are estranged from each other. Finally, we propose a domain augmentation strategy that can randomly generate new domains for model generalization training. Experimental results on multi-site fundus image datasets for optic cup and disc segmentation show that the CDD has good model generalization. Our proposed CDD outperforms several state-of-the-art methods in domain generalizable segmentation.

\end{abstract}
\section{Introduction}
Deep learning with Convolutional Neural Networks (CNNs) have achieved remarkable performance in medical image segmentation~\cite{litjens2014evaluation,ronneberger2015u,shen2017deep}. Learning discriminative features for a given task is crucial for the segmentation accuracy and robustness. Recently, the feature discrimination can be achieved by disentangling mutual information between latent variables and data variations~\cite{chen2016infogan}. However, it can only achieve high-quality disentanglement
and reconstruction on samples in seen domains while cannot perform well on unseen domains, which is often called Domain Generalization (DG) problem. Hence, it is desirable to improve disentangle models to possess model performance and generalization.


The idea of disentangling features has been employed to learn domain-invariant features in multi-modality datasets. They are usually adopted for dealing with Domain Adaptation (DA) problem. Yang et al.~\cite{yang2019unsupervised} disentangled the images from each domain into a shared domain-invariant content space and a domain-specific style space, then used the representation in the content space to perform liver segmentation. Pei et al.~\cite{pei2021disentangle} implemented domain adaptation on both feature and image levels with disentangled domain-invariant and domain-specific features and they further introduced zero loss to encourage a more thorough disentanglement. However, these disentangle-based unsupervised DA methods separately distribute encoder and decoder to each domain leading the model to be adapted to one domain, and they require target samples for reference in training, which is impractical because it is impossible to specially re-train an exclusive model adapting to a domain set. Therefore, it is desirable to enable the model to capture shared content representations across multi-domains under an union disentangle structure. Chartsias et al.~\cite{chartsias2019disentangled} proposed Spatial Decomposition Network (SDNet) that can decompose the input image into a spatial anatomical factor and a non-spatial modality factor. SDNet offers only a single encoder for disentanglement, holding the potential to extract general anatomical representation useful for multi-task learning. Although SDNet demonstrates that it is suitable for multi-modality learning to use an union disentangled structure, it can only perform disentanglement and reconstruction well on seen domains while performs poorly on unseen domains that are not involved in training. The reason is mainly because of the unconstrained distribution of style codes without sufficient discrimination across multiple sites hampering the extraction of domain-invariant feature representation.

Recently, developing model generalization has been brought into focus and applied on medical imaging~\cite{dou2019domain}. 
Wang et al.~\cite{wang2020dofe} proposed DoFE to dynamically enrich the image features with domain prior knowledge learned from multi-source domains to make the semantic features more discriminative. Gu et al.~\cite{gu2021domain} proposed a domain composition method that dynamically represent domain features by a linear combination of basis representations. These methods share the main idea of using domain-discriminative representations to simulate a specific domain.
Contrastively, disentanglement offers a more thorough strategy to decompose domain feature into domain-invariant contents and domain-specific styles. How to efficiently use the domain-invariant and domain-specific information is an important problem to solve for model generalization. Hence, it prompts us to explore the potential of conducting generalizable medical image segmentation with disentangle model to efficiently utilize domain-invariant contents.

In this work, we propose a Contrastive Domain Disentangle (CDD) network for generalizable medical image segmentation. First, we introduce a disentangled representation learning strategy to jointly decompose each image into anatomical representation and modality representation. We further use the decomposed anatomical representation to achieve generalizable medical image segmentation. The decomposition and reconstruction branches in the disentangle model guarantee it to extract independent anatomical and style representations. Second, we propose a domain style contrasitve loss to encourage the decomposed modality factors to have a consistent distribution with the same domain and divergent distribution between different domains. Finally, we propose a domain-augmentation strategy to generate various domain datasets for model training. Comprehensive experimental results on multi-site fundus images show that our proposed contrastive disentangle-based method outperforms several state-of-the-art domain generalization methods.

\section{Methods}
\begin{figure}
    \centering
    \includegraphics[width=\textwidth]{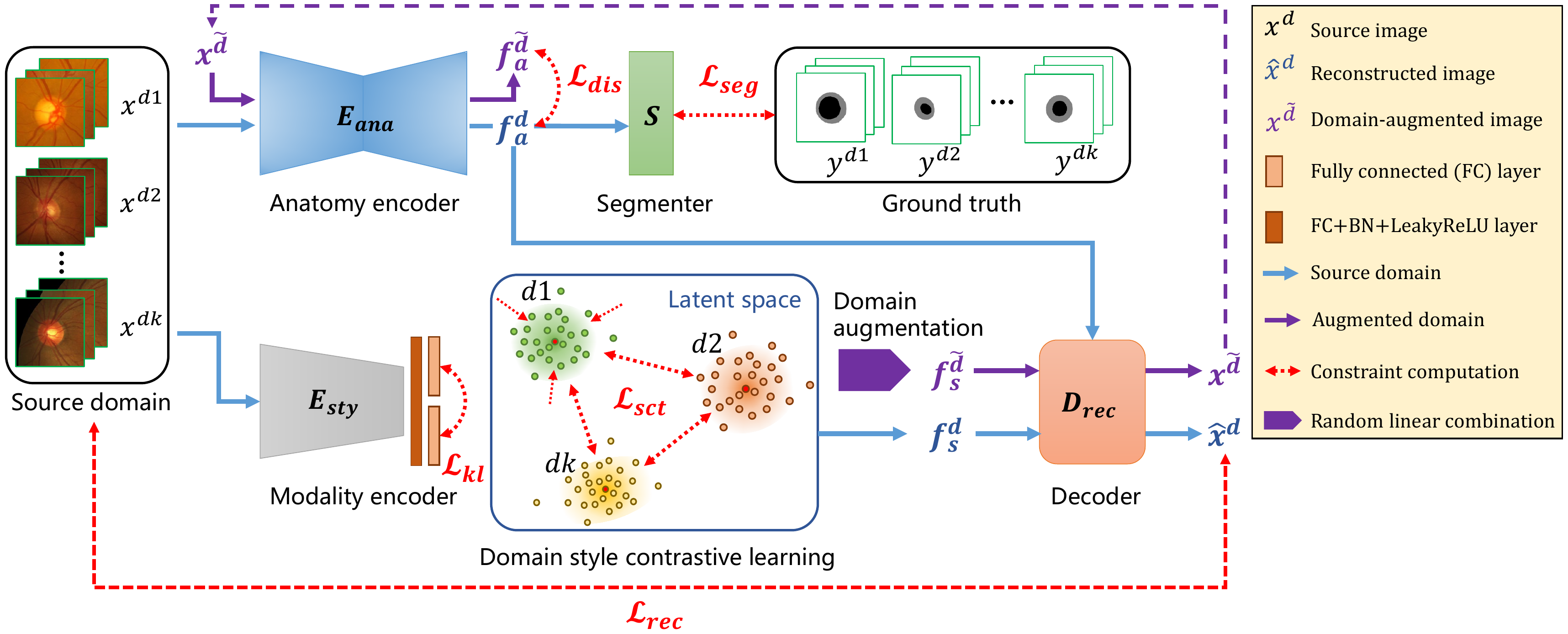}
    \caption{Overview of the proposed Contrastive Domain Disentangle (CDD) network for multi-site domain generalizable segmentation. CDD contains an anatomy encoder $E_{ana}$, a style encoder $E_{sty}$ and a reconstruction decoder $D_{rec}$. $E_{ana}$ and $E_{sty}$ extract anatomical representations $f_{a}^{d}$ and modality representations $f_{s}^{d}$, respectively. $D_{rec}$ takes $f_{a}^{d}$ and $f_{s}^{d}$ as input and obtains a reconstruction of the input image $\hat{x}^{d}$. $x^{\tilde{d}}$ is our simulated image using domain-augmentation. The decomposed anatomical representations are further used for segmentation.}
    \label{fig1:model}
\end{figure}

Our proposed Contrastive Domain Disentanglement (CDD) network is illustrated in Fig.~\ref{fig1:model}. It contains an anatomy encoder $E_{ana}$, a style encoder $E_{sty}$ and a reconstruction decoder $D_{rec}$ that takes a combination of an anatomical representation and a style representation to reconstruct an image. Supposing a multi-domain training dataset $\mathcal{D}=\{x_i^d,y_i^d\}_{i=1}^{N_d}, d={d_1,d_2,...,d_k}$ is collected from different domains (e.g., with different imaging devices), where $x_i^d$ depicts the $i$th training sample from the $d$th source domain with its corresponding ground-truth annotation $y_i^d$. $N_d$ denotes the number of training samples in domain $d$. 
\subsubsection{Domain Disentangle Network.} For disentanglement, each input image $x_{i}^{d}\in \mathbb{R}^{H\times W\times C}$ passes through $E_{ana}$ to get the domain-invariant anatomical representations $f_{i,a}^d\in f_{a}^d:= \{0,1\}^{H\times W\times T}$, where $T$ is the number of channels for domain-invariant representation. $H$ and $W$ correspond to the height and width of the image~\cite{chartsias2019disentangled}. Meanwhile, $x_{i}^{d}$ passes through $E_{sty}$ to obtain the modality representations $f_{i,s}^d\in f_{s}^d:=\mathbb{R}^{1\times Z}$, where $Z$ means the number of factors in the representations. Finally, the $f_{i,a}^d$ and $f_{i,s}^d$ are combined by a decoder $D_{rec}$ to gain the reconstructed source-like images $\hat{x}_{i}^d$.

The decomposed anatomical representation $f_{a}^d$ is further fed into a segmentation network $S$ for supervised training. 
Here, we employ the classical U-Net~\cite{ronneberger2015u} with Bath Normalization (BN) as the backbone in $E_{ana}$. The $f_{a}^d$ is derived using a gumbel softmax~\cite{jang2016categorical} activation function to binarize the anatomical representation while guarantee the gradient back propagation. Taking domain $d$ as instance, the supervised segmentation training is computed as:
\begin{equation}
\small
\label{eq1:loss_seg}
    \mathcal{L}_{seg} = \frac{1}{2N_{d}}\sum_{i=1}^{N_{d}}\big( \mathcal{L}_{Dice}(p_i^{d},y_i^{d}) + \mathcal{L}_{ce}(p_i^{d},y_i^{d})\big)
\end{equation}
where $p_i^d$ means the prediction of $i$th image in domain $d$. Here, we use a hybrid segmentation loss that consists of the cross-entropy loss $\mathcal{L}_{ce}$ and Dice loss $\mathcal{L}_{Dice}$.

$E_{sty}$ extracts modality representations based on a Variational Autoencoder (VAE)~\cite{kingma2013auto}. The VAE learns a low dimensional latent space so that the learned latent representations match a prior distribution that is set to an isotropic multivariate Gaussian $p(z)=\mathcal{N}(0, 1)$. Given the input $x^d$, $E_{sty}$ estimates Gaussian distribution with two output parameters (the mean and the variance of the distribution, $f_{mea}^d$ and $f_{var}^d$). This distribution is then reparameterized to obtain the modality style representation $f_{s}^{d}$. $E_{sty}$ and $D_{rec}$ are trained by minimising a reconstruction error $\mathcal{L}_{rec}$ and KL divergence is computed between the estimated Gaussian distribution $q(z|f_{mea}^d, f_{var}^d)$ and the unit Gaussian $p(z)$:
\begin{equation}
\small
\label{eq2:loss_rec}
    \mathcal{L}_{rec} = \frac{1}{N_{d}}\sum_{i=1}^{N_{d}}\big|x_i^d-\hat{x}_i^d\big|
\end{equation}
\begin{equation}
\small
\label{eq3:loss_kl}
    \mathcal{L}_{kl} = D_{kl}\big(q(z|f_{mea}^d,f_{var}^d) \| p(z)\big)
\end{equation}
where $D_{kl}(p\|q)=\sum p(x)log\frac{p(x)}{q(x)}$. When $E_{sty}$ is trained, sampling a vector from the unit Gaussian and passing it through the $D_{rec}$ approximates sampling from the corresponding domain. 
\subsubsection{Domain Style Contrastive Learning.}
In order to encourage model decomposed the more discriminative style representation. A style contrastive loss function is proposed to encourage the style representations from the same domain to be close and those from different domains to be divergent from each other. At an iteration, we randomly sample a mini-batch set of representations in each domain $d$ with size $b$ which is demonstrated as $f_{b,s}^d$. Specifically, given the mini-batch vectors from domain $d$ coupled with its randomly permuted sample $\acute{f}_{b,s}^d$, we obtain a positive pair $\{f_{b,s}^d,\acute{f}_{b,s}^d\}$. Meanwhile, we set the negative pairs as $\{f_{b,s}^d, f_{b,s}^{\hat{d}}\}$, where $\hat{d}$ means a domain other than $d$. Following the standard formula of self-supervised contrastive loss~\cite{chen2020simple,wang2021dense}, we define a domain style contrastive loss as:
\begin{equation}
\label{eq4:loss_sct}
\small
    \mathcal{L}_{sct} = \frac{1}{|b|}\sum_{i=1}^{b}-log\frac{e^{sim(f_{i,s}^d, \acute{f}_{i,s}^d)/\tau}}{e^{sim(f_{i,s}^d, \acute{f}_{i,s}^d)/\tau}+\sum_{j\in\hat{d}} e^{sim(f_{i,s}^d, f_{i,s}^j)/\tau}}
\end{equation}
where the $sim(\cdot,\cdot)$ is the cosine similarity between two representations, and $\tau=0.1$ is the temperature scaling parameter.
\subsubsection{Domain-Augmentation Strategy}
To let the model be generalizable to unseen target domain, we propose a domain-augmentation strategy to automatically simulate domain representations $f_s^{\tilde{d}}$. The domain augmentation obtains a new modality representation $f_s^{\tilde{d}}$ based on a linear combination of source modality representations $f_s^{d}$. Hence, when finishing disentanglement, we generate random weight parameters $\alpha^d = [-1, 1]$ corresponding to each domain $d$. Smoothly, the new simulated modality representations are computed as $f_s^{\tilde{d}} = \sum_{i=d_{1}}^{d_{k}}\alpha^{i}f_s^i$. Finally, the new simulated domain dataset $x^{\tilde{d}}$ can be reconstructed by $D_{rec}$ with the combination of $f_s^{\tilde{d}}$ and $f_a^d$. To encourage the $E_{ana}$ to perform a thorough decomposition ability and be robust on unseen-domain images, we feed $x^{\tilde{d}}$ into $E_{ana}$ and obtain their corresponding anatomical representations $f_a^{\tilde{d}}$. We introduce L1 distance to constrain the consistence between $f_a^{\tilde{d}}$ and $f_a^{d}$:
\begin{equation}
\small
\label{eq5:loss_dis}
    \mathcal{L}_{dis} = \frac{1}{ N_{d}}\sum_{i=1}^{N_{d}}\big|f_{i,a}^d-f_{i,a}^{\tilde{d}}\big|
\end{equation}
\subsubsection{Overall Loss.}
The overall loss for training is a combination of the segmentation loss $\mathcal{L}_{seg}$, the reconstruction loss $\mathcal{L}_{rec}$,
the VAE loss $\mathcal{L}_{kl}$, the domain style contrastive loss $\mathcal{L}_{sct}$ and the anatomical representation consistence loss $\mathcal{L}_{dis}$:
\begin{equation}
\small
\label{eq6:loss_all}
    \mathcal{L} = \mathcal{L}_{seg} + \lambda_{1}\mathcal{L}_{rec} + \lambda_{2}\mathcal{L}_{kl} + 
    \lambda_{3}\mathcal{L}_{sct} +
    \lambda_{4}\mathcal{L}_{dis}
\end{equation}
where $\lambda_1,\lambda_2,\lambda_3$ and $\lambda_4$ act as trade-off parameters with each loss function.
\section{Experiments and Results}
\subsubsection{Datasets and Implementation.} For experiment, we evaluate our approach for Optic Cup (OC) and Disc (OD) segmentation on the well-organised multi-site retinal fundus images\footnote{https://github.com/emma-sjwang/Dofe} with image size $384\times384$~\cite{wang2020dofe}. The multi-site datasets are collected from four public fundus image datasets imaged with different scanners in different sites that have distinct domain discrepancies in visualization and image quality: Domain 1 is from Drishti-GS~\cite{sivaswamy2015comprehensive} dataset containing 50 and 51 images for training and testing, respectively; Domain 2 is from RIM-ONE~\cite{fumero2011rim} dataset containing 99 and 60 images for training and testing, respectively; Domain 3 and 4 are respectively from REFUGE~\cite{orlando2020refuge} challenge's train and val datesets both containing 320 and 80 images for training and testing. For assessing the generalizable OC$/$OD segmentation in unseen-domain, we follow the leave-one-domain-out strategy in DoFE~\cite{liu2020shape}, where each time one domain dataset is used as target domain to evaluate the generalizable performance and the remaining three datasets are used as source domains for network training.

The $E_{ana}$ employs classical U-Net as the backbone with the channel numbers of 16, 32, 64, 128 and 256 at five scales, respectively~\cite{ronneberger2015u}. We set the channel of output anatomical representations $T$ as 8. The segmenter $S$ contains of 2 convolutional blocks including convolution layer with respective kernel size $3\times3$ and $1\times1$, BN and LeakyReLU (sloop $=0.2$) layers, followed by a convolution layer to output the predictions. The $E_{style}$ is composed with 4 blocks of convolutional blocks followed by 2 fully connected layers to output the two parameters of Gausian estimation ($f_{mea}$ and $f_{var}$). We set the channel of output modality representations $Z$ as 16. The loss function weights \{$\lambda_{1}$, $\lambda_{2}$, $\lambda_{3}$, $\lambda_{4}$\} are set as \{1, 0.001, 0.01, 1\}, respectively. For training, we adopt a series of basic data augmentation to enhance the diversity of training samples as conducted by DoFE~\cite{wang2020dofe}. The input image is obtained by random cropping with the size of $256\times256$. 
The proposed disentangle network is trained by an end-to-end manner using Adam optimizer and the initial learning rate was $1e^{-3}$. The learning rate will decay to 95\% when the model does not improve in 8 epoches. We trained 200 epoches with batch size of 10 and the mini-batch size $b$ was 8 in the each domain. Training was implemented on one NVIDIA Geforce GTX 1080 Ti GPU. For fair comparison, we kept most parameters be same as those in DoFE~\cite{wang2020dofe}. We adopt Dice score (Dice) and Average Surface Distance (ASD) for evaluation.
\begin{table}
    \centering
    \caption{Generalizable OC$/$OD segmentation performance in Dice (\%) on multi-site fundus image datasets of various methods. CDD$_{base}$ means the modified disentangle baseline. CDD$_{sct}$ means baseline network introduces style contrastive learning. CDD$_{da}$ means baseline network introduces domain augmentation strategy. CDD$_{sctda}$ means baseline network introduces both style contrastive learning and domain augmentation strategy.} \label{tab1:dice}
    \scalebox{0.67}{\begin{tabular}{l|c|c|c|c|c|c|c|c|c}
    \hline
    \hline
    \multirow{2}{0.74in}{Methods} & \multicolumn{2}{c}{Domain 1} & \multicolumn{2}{|c|}{Domain 2} & \multicolumn{2}{|c|}{Domain 3} & \multicolumn{2}{|c|}{Domain 4} &
    \multirow{2}{*}{Avg} \\ \cline{2-9}
    & cup & disc & cup & disc & cup & disc & cup & disc & \\ \hline
    Inter-domain & 74.38$\pm$12.96 & 96.67$\pm$2.04 & 77.71$\pm$20.84 & 85.05$\pm$14.67 & 79.72$\pm$9.51 & 90.01$\pm$5.81 & 86.63$\pm$8.52 & 89.55$\pm$3.26 & 84.97 \\
    Intra-domain & 83.35$\pm$13.99 & 96.10$\pm$1.88 & 81.53$\pm$9.42 & 94.62$\pm$3.01 & 87.57$\pm$7.59 & 95.91$\pm$1.85 & 88.88$\pm$7.10 & 95.58$\pm$1.98 & 90.44 \\ \hline
    BigAug~\cite{zhang2020generalizing} & 82.36$\pm$11.74 & 93.73$\pm$9.29 & 75.45$\pm$15.01 & 87.83$\pm$11.17 &
    84.32$\pm$9.45 & 91.99$\pm$10.72 &
    85.32$\pm$7.50 & 92.97$\pm$6.58 & 86.87 \\
    DoFE~\cite{wang2020dofe} & 80.25$\pm$10.84 & 95.61$\pm$1.45 & 78.97$\pm$14.80 & 88.74$\pm$4.58 &
    84.81$\pm$7.71 & 92.81$\pm$2.63 & 86.65$\pm$6.39 & 93.46$\pm$2.43 & 87.66 \\ 
    DCAC~\cite{hu2021domain} & 82.09$\pm$13.71 & 95.46$\pm$8.93 & 77.72$\pm$18.96 & 87.20$\pm$11.77 & 86.12$\pm$7.61 & 92.50$\pm$7.15 & 86.55$\pm$9.30 & 94.07$\pm$8.42 & 87.71 \\ \hline
    CDD$_{base}$ & 80.63$\pm$11.55 & 95.02$\pm$2.65 &
    79.35$\pm$13.66 & \textbf{89.76$\pm$3.20} &
    83.29$\pm$8.04 & \textbf{93.67$\pm$3.36} & 84.12$\pm$11.33 & 93.51$\pm$4.03 & 87.42 \\
    CDD$_{sct}$ & 80.64$\pm$11.94 & 96.11$\pm$2.95 &
    80.13$\pm$17.39 & 88.27$\pm$1.23 &
    85.20$\pm$8.06 & 92.36$\pm$2.59 & 86.33$\pm$9.12 & 93.36$\pm$2.60 & 87.80 \\
    CDD$_{da}$ & 84.28$\pm$11.56 & 96.13$\pm$1.35 &
    81.95$\pm$12.60 & 88.18$\pm$3.50 &
    84.59$\pm$8.11 & 92.95$\pm$3.30 & 86.49$\pm$9.69 & 93.27$\pm$3.40 & 88.48 \\
    CDD$_{sctda}$ & \textbf{85.75$\pm$12.31} & \textbf{96.79$\pm$1.53} & \textbf{81.04$\pm$13.63} & 89.71$\pm$3.60 &  \textbf{86.94$\pm$7.94} & 93.25$\pm$3.55 & \textbf{86.86$\pm$8.97} & \textbf{94.44$\pm$3.96} & \textbf{89.35} \\
    \hline
    \hline
    \end{tabular}}
\end{table}
\begin{table}
    \centering
    \caption{Generalizable OC$/$OD segmentation performance in ASD (pixel) on multi-site fundus image datasets of various methods.}\label{tab2:asd}
    \scalebox{0.68}{\begin{tabular}{l|c|c|c|c|c|c|c|c|c|c}
    \hline
    \hline
    \multirow{2}{0.9in}{Methods} & \multicolumn{2}{c}{Domain 1} & \multicolumn{2}{|c|}{Domain 2} & \multicolumn{2}{|c|}{Domain 3} & \multicolumn{2}{|c|}{Domain 4} & \multicolumn{2}{|c}{Avg} \\ \cline{2-11}
    & cup & disc & cup & disc & cup & disc & cup & disc & cup & disc \\ \hline
    Inter-domain & 22.35$\pm$9.74 & 6.47$\pm$3.80 & 15.77$\pm$20.21 & 18.25$\pm$19.60 & 12.30$\pm$5.82 & 12.33$\pm$5.03 & 7.45$\pm$4.60 & 9.27$\pm$2.62 & 14.47 & 11.58 \\
    Intra-domain & 16.04$\pm$6.65 & 7.84$\pm$3.87 & 13.10$\pm$7.68 & 8.55$\pm$5.80 & 8.41$\pm$5.02 & 6.32$\pm$4.02 & 6.07$\pm$3.41 & 5.46$\pm$2.48 & 10.91 & 7.04 \\ \hline
    BigAug~\cite{zhang2020generalizing} & 17.91$\pm$10.11 & 8.67$\pm$4.08 & 22.33$\pm$15.26 & 19.77$\pm$6.69 &
    13.51$\pm$7.67 & 14.46$\pm$4.96 &
    8.90$\pm$5.02 & 8.77$\pm$6.63 & 15.66 & 12.92 \\
    DoFE~\cite{wang2020dofe} & 17.16$\pm$9.40 & 7.62$\pm$2.38 & 15.28$\pm$12.94 & 14.52$\pm$5.36 &
    10.73$\pm$6.22 & 10.11$\pm$5.11 & \textbf{7.18$\pm$3.23} & 7.60$\pm$3.64 & 12.59 & 9.96 \\ 
    DCAC~\cite{hu2021domain} & 18.97$\pm$11.83 & 7.83$\pm$3.11 & 16.98$\pm$12.75 & 14.15$\pm$5.17 & 10.25$\pm$6.20 & 9.64$\pm$3.42 & 8.87$\pm$4.18 & \textbf{6.47$\pm$3.52} & 13.77 & 9.92 \\ \hline
    CDD$_{base}$ & 17.33$\pm$8.58 & 8.21$\pm$4.61 &
    13.10$\pm$6.66 & \textbf{11.79$\pm$3.90} & 11.03$\pm$5.22 & \textbf{9.31$\pm$4.23} & 8.04$\pm$6.42 & 7.43$\pm$4.89 & 12.38 & 9.18 \\
    CDD$_{sct}$ & 18.21$\pm$8.05 & 7.52$\pm$5.85 &
    13.33$\pm$13.43 & 14.10$\pm$13.50 &
    10.09$\pm$5.42 & 9.98$\pm$3.11 & 7.30$\pm$3.88 & 7.03$\pm$2.57 & 12.23 & 9.66 \\
    CDD$_{da}$ & 15.77$\pm$6.93 & 7.14$\pm$2.59 &
    \textbf{11.04$\pm$5.91} & 12.97$\pm$3.94 &
    10.58$\pm$5.21 & 9.60$\pm$3.71 & 7.22$\pm$5.31 & 7.51$\pm$4.10 & 11.15 & 9.31 \\
    CDD$_{sctda}$ & \textbf{14.65$\pm$8.39} & \textbf{6.54$\pm$3.74} & 12.91$\pm$10.79 & 13.06$\pm$8.60 &  \textbf{9.38$\pm$5.40} & 9.32$\pm$4.11 & 7.28$\pm$5.85 & 6.87$\pm$5.03 & \textbf{11.06} & \textbf{8.95} \\
    \hline
    \hline
    \end{tabular}}
\end{table}
\subsubsection{Generalizable Fundus Image Segmentation.} We used leave-one-domain-out for the DG experiment. We first considered all the available training domains as a single dataset and trained a U-Net based on a standard Dice loss without considering the domain difference and directly apply it to th unseen domain, which is referred to as 'Inter-domain' and serves as a lower bound of the experiment. Then we consider training and testing the segmentation model for each domain respectively, which serves as the upper bound for DG and referred to as 'Intra-domain'. For DG methods, we compared our proposed Contrastive Domain Disentangle (CDD) network with three representative state-of-the-art generalization methods, including the data-augmentation based method (BigAug)~\cite{li2018domain}, the Domain-oriented Feature Embedding based method  (DoFE)~\cite{wang2020dofe}, and the Domain and Content Adaptive Convolution
based method (DCAC)~\cite{hu2021domain}.

\begin{figure}
    \centering
    \includegraphics[width=0.8\textwidth]{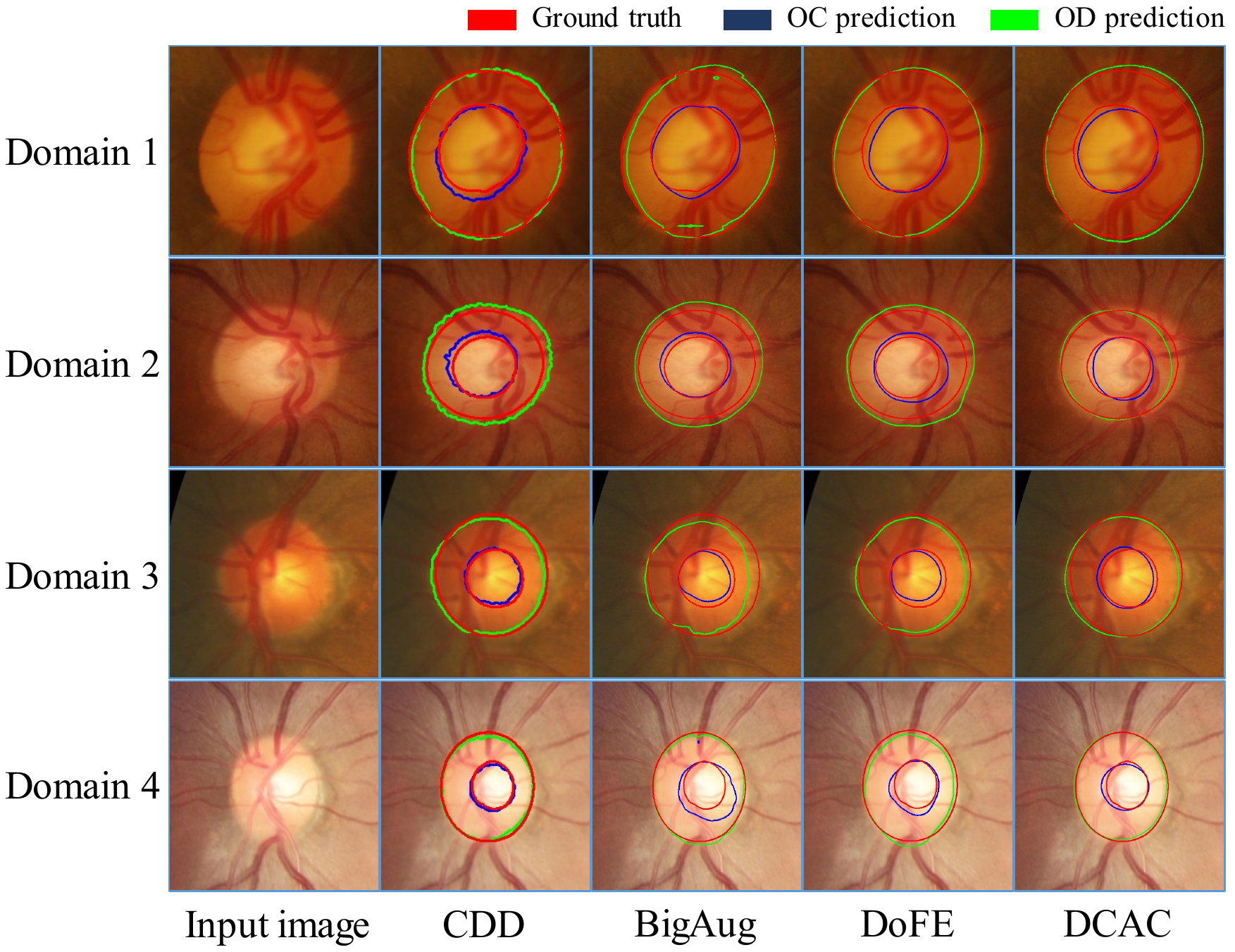}
    \caption{Visual comparison between our proposed CDD and BigAug~\cite{zhang2020generalizing}, DoFE~\cite{wang2020dofe}, DCAC~\cite{hu2021domain}.} 
    \label{fig2:results}
\end{figure}

Table~\ref{tab1:dice} and Table~\ref{tab2:asd} show the quantitative evaluation results in OC$/$OD generalizable segmentation. Intra-domain achieves the highest performance both in Dice score and ASD evaluations, getting 90.44$\%$ of Dice, 10.91 pixels in OC and 7.04 pixels in OD of ASD, respectively. The performance gap between lower and upper bound is about 5$\%$ in Dice score. BigAug obtains a slight improvement from Inter-domain, which suggests that aimlessly conducting data-augmentation owns greater uncertainty for improving the model generalization. In contrast, DoFE and DCAC perform better than BigAug, but they just implicitly utilize these domain discriminative representations and do not introduce domain constraints to obtain a more thorough domain representations. In comparison, our proposed CDD network makes further improvements achieving the highest Dice score and the best ASD compared with the other methods. The results from our proposed method are closer to the upper-bound. Fig.~\ref{fig2:results} provides the visual comparison between our proposed Contrastive Domain Disentangle (CDD) network and BigAug, DoFE as well as DCAC for images from the 4 domains, respectively. The visualizations show that the predictions of our proposed CDD achieve closer prediction boundary to the ground truth. While, the other domain generalization methods have more over- and mis-predicted regions than ours.
\subsubsection{Ablation Study.}
We proved the effectiveness of our proposed domain style contrastive learning and domain augmentation strategy. CDD$_{base}$ is actually SDNet~\cite{chartsias2019disentangled} for feature disentangle only and serves as a baseline. CDD$_{sct}$ means baseline $+$ domain style contrastive learning; CDD$_{da}$ means baseline $+$ domain augmentation strategy and CDD$_{sctda}$ means CDD$_{sct}$ $+$ domain augmentation. 
As the results shown in Table.~\ref{tab1:dice}, CDD$_{base}$ obtained an average Dice of 87.42$\%$, and using domain style contrastive learning improved it to 87.80$\%$, indicating that encouraging the network to decompose into more representative domain feature benefits for multi-modality learning. Then, CDD$_{da}$ got the average Dice of 88.48$\%$, proving that domain augmentation is effective for improving model generalization. Finally, combining them together will further improved it to 89.35$\%$, indicating that our proposed CDD with domain style contrastive learning and domain augmentation strategy building the comprehensive model generalization.
\section{Conclusion}
We presented a Contrastive Domain Disentangle network (CDD) to tackle the domain generalization problem in medical image segmentation. We introduced a disentangle network for decomposing medical images into an anatomical representation factor and a modality representation factor. We can efficiently model the domain style variability of medical data by the incorporation of a variational autoencoder (VAE). The VAE model can also reconstruct the images using the combination of the two representation factors. Then, a style contrastive learning loss is proposed to encourage the modality representations from the sample domain to distribute as close as possible while different domains are estranged from each other. Finally, we proposed a domain augmentation strategy that can randomly generate domains other than the source domains for model generalization training. Experimental results showed the effectiveness of our proposed CDD for achieving robust performance in OC$/$OD segmentation from multi-site fundus image datasets. In the future, it is of interest to apply our proposed CDD network to other multi-modality medical image datasets to validate its powerful model generalization.

%
%
%
\bibliographystyle{splncs04}
\bibliography{references}
\end{document}